\pdfoutput=1
\documentclass[fleqn,10pt]{wlscirep}
\usepackage[utf8]{inputenc}
\usepackage[T1]{fontenc}
\usepackage{lineno}

\title{A dataset for plain language adaptation of biomedical abstracts}

\author[1,*]{Kush Attal}
\author[1]{Brian Ondov}
\author[1]{Dina Demner-Fushman}
\affil[1]{Lister Hill National Center for Biomedical Communications, U.S. National Library of Medicine, National Institutes of Health, Bethesda, MD, USA }

\affil[*]{corresponding author: Kush Attal (Kush.Attal@nih.gov)}

\begin{abstract}
Though exponentially growing health-related literature has been made available to a broad audience online, the language of scientific articles can be difficult for the general public to understand. Therefore, adapting this expert-level language into plain language versions is necessary for the public to reliably comprehend the vast health-related literature. Deep Learning algorithms for automatic adaptation are a possible solution; however, gold standard datasets are needed for proper evaluation. Proposed datasets thus far consist of either pairs of comparable professional- and general public-facing documents or pairs of semantically similar sentences mined from such documents. This leads to a trade-off between imperfect alignments and small test sets. To address this issue, we created the Plain Language Adaptation of Biomedical Abstracts dataset. This dataset is the first manually adapted dataset that is both document- and sentence-aligned. The dataset contains 750 adapted abstracts, totaling 7643 sentence pairs. Along with describing the dataset, we benchmark automatic adaptation on the dataset with state-of-the-art Deep Learning approaches, setting baselines for future research.
\end{abstract}
\begin{document}

\flushbottom
\maketitle

\thispagestyle{empty}

\section*{Background \& Summary}

While reliable resources for health information conveyed in a plain language format exist, such as the MedlinePlus website from the National Library of Medicine (NLM) \cite{noauthor_medlineplus_nodate}, these resources do not provide all the necessary information for every health-related situation or rapidly changing state of knowledge arising from novel scientific investigations or global events like pandemics. 
In addition, the language used in other health-related articles can be too difficult for patients and the general public to comprehend\cite{ROSENBERG201757}, which has a major impact on health outcomes\cite{stableford2007plain}. While work in simplifying text exists, the unique language of biomedical text warrants a distinct subtask similar to machine translation, termed adaptation\cite{xu_optimizing_2016}. Adapting natural language involves creating a simplified version that maintains the most important details from a complex source. Adaptations are a common tool for teachers to use to improve comprehension of content for English language learners\cite{carlo2004closing}.  

A standard internet search will return multiple scientific articles that correspond to a patient’s query; however, without extensive clinical and/or biological knowledge, the user may not be able to comprehend the scientific language and content\cite{white_cyberchondria_2009}. There are articles with verified, plain language summaries for health information, such as the articles with corresponding plain language summaries created by medical health organization Cochrane\cite{noauthor_cochrane_nodate}. However, creating manual summaries and adaptations for every article addressing every user's queries is not possible. Thus, an automatic adaptation generated for material responding to a user’s query is very relevant, especially for patients without clinical knowledge. 


Though plain language thesauri and other knowledge bases have enabled rule-based systems that substitute difficult terms for more common ones, human editing is needed to account for grammar, context, and ambiguity\cite{kauchak_web-based_2020}. Deep Learning may offer a solution for fully automated adaptation. 
Advances in architectures, hardware, and available data have led neural methods to achieve state-of-the-art results in many linguistic tasks, including Machine Translation\cite{stahlberg2020neural} and Text Simplification\cite{al2021automated}. Neural methods, however, require large numbers of training examples, as well as benchmark datasets to allow iterative progress\cite{savery2020question}.



Parallel datasets for Text Simplification have been assembled by searching for semantically similar sentences across comparable document pairs, for example articles on the same subject in both Wikipedia and Simple English Wikipedia (or Vikidia, an encyclopedia for children in several languages)\cite{jiang_neural_2020,coster_simple_2011,hwang_aligning_2015,zhu_monolingual_2010}. Since Wikipedia contains some articles on biomedical topics, it has been proposed to extract subsets of these datasets for use in this domain\cite{van_automets_2020,van_den_bercken_evaluating_2019, adduru_towards_2018, cardon_parallel_2019}. However, since these sentence pairs exist in different contexts, they are often not semantically identical, having undergone sentence-level operations like splitting or merging. Sentence pairs pulled out of context may also use anaphora on one side of a pair but not the other. This can confuse models during training and expect impossible replacements during testing. Further, Simple English Wikipedia often still contains complex medical terms on the simple side\cite{van_automets_2020,xu_problems_2015,shardlow-nawaz-2019-neural}. Parallel sentences have also been mined from dedicated biomedical sources. Cao et al. have expert annotators pinpoint highly similar passages, usually consisting of one or two sentences from each passage, from Merck Manuals, an online website containing numerous articles on medical and health topics created for both professional and general public groups\cite{cao_expertise_2020}. In addition, Pattisapu et al. have expert annotators identify highly similar pairs from scientific articles and corresponding health blogs describing them\cite{pattisapu_leveraging_2020}. Though human filtering makes the pairs in both these datasets much closer to being semantically identical, at less than 1,000 pairs each, they are too small for training and even less than ideal for evaluation\cite{vstajner2022sentence}. Sakakini et al. manually translate a somewhat larger set (4,554) of instructions for patients from clinical notes\cite{sakakini_context-aware_2020}. However, this corpus covers a very specific case within the clinical domain, which itself constitutes a separate sublanguage from biomedical literature\cite{friedman_two_2002}.

Since recent models can handle larger paragraphs, comparable corpora have also been suggested as training or benchmark datasets for adapting biomedical text. These corpora consist of pairs of paragraphs or documents that are on the same topic and make roughly the same points, but are not sentence-aligned.
Devaraj et al. present a paragraph level corpus derived from Cochrane review abstracts and their Plain Language Summaries, using heuristics to combine subsections with similar content across the pairs. However, these heuristics do not guarantee identical content\cite{basu2021automatic}. This dataset is also not sentence-aligned, which limits the architectures that can take advantage of it and results in restriction of documents to those with no more than 1024 tokens.
Other datasets include comparable corpora or are created at the paragraph-level and omit relevant details from the original article\cite{basu2021automatic}.
To the best of our knowledge, no datasets provide manual, sentence-level adaptations of the  scientific abstracts\cite{ondov2022survey}.
Thus, there is still a need for a high-quality, sentence-level gold standard dataset for the adaptation of general biomedical text.

To address this need, we have developed the Plain Language Adaptation of Biomedical Abstracts (PLABA) dataset. PLABA contains 750 abstracts from PubMed (10 on each of 75 topics) and expert-created adaptations at the sentence-level. Annotators were chosen from the NLM and an external company and given abstracts within their respective expertise to adapt.
Human adaptation allows us to ensure the parallel nature of the corpus down to sentence-level granularity, but still while using the surrounding context of the entire document to guide each translation. We deliberately construct this dataset so it can serve as a gold standard on several levels:
\begin{enumerate}
    \item 
Document level simplification. Documents are simplified in total, each by at least one annotator, who is instructed to carry over all content relevant for general public understanding of the professional document. This allows the corpus to be used as a gold standard for systems that operate at the document level.
    \item
Sentence level simplification. Unlike automatic alignments, these pairings are ensured to be parallel for the purpose of simplification. Semantically, they will differ only in (1) content removed from the professional register because the annotator deemed it unimportant for general public understanding, and (2) explanation or elaboration added to the general public register to aid understanding. Since annotators were instructed to keep content within sentence boundaries (or in split sentences), there are no issues with fragments of other thoughts spilled over from neighboring sentences on one side of the pair.
    \item
Sentence-level operations and splitting. Though rare in translation between languages, sentence-level operations (e.g. merging, deletion, and splitting) are common in simplification\cite{frankenberg-garcia_corpus_2019}. Splitting is often used to simplify syntax and reduce sentence length. Occasionally sentences may be dropped from the general public register altogether (deletion). For consistency and simplicity of annotation, we do not allow merging, creating a one-to-many relationship at the sentence level.
\end{enumerate}

The PLABA dataset should further enable the development of systems that automatically adapt relevant medical texts for patients without prior medical knowledge. In addition to releasing PLABA, we have evaluated state-of-the-art deep learning approaches on this dataset to set benchmarks for future researchers. 

\section*{Methods}

The PLABA dataset includes 75 health-related questions asked by MedlinePlus users, 750 PubMed abstracts from relevant scientific articles, and corresponding human created adaptations of the abstracts. The questions in PLABA are among the most popular topics from MedlinePlus, ranging from topics like COVID-19 symptoms to genetic conditions like cystic fibrosis\cite{noauthor_medlineplus_nodate}.

\begin{figure}[ht]
\centering
\includegraphics[width=\textwidth]{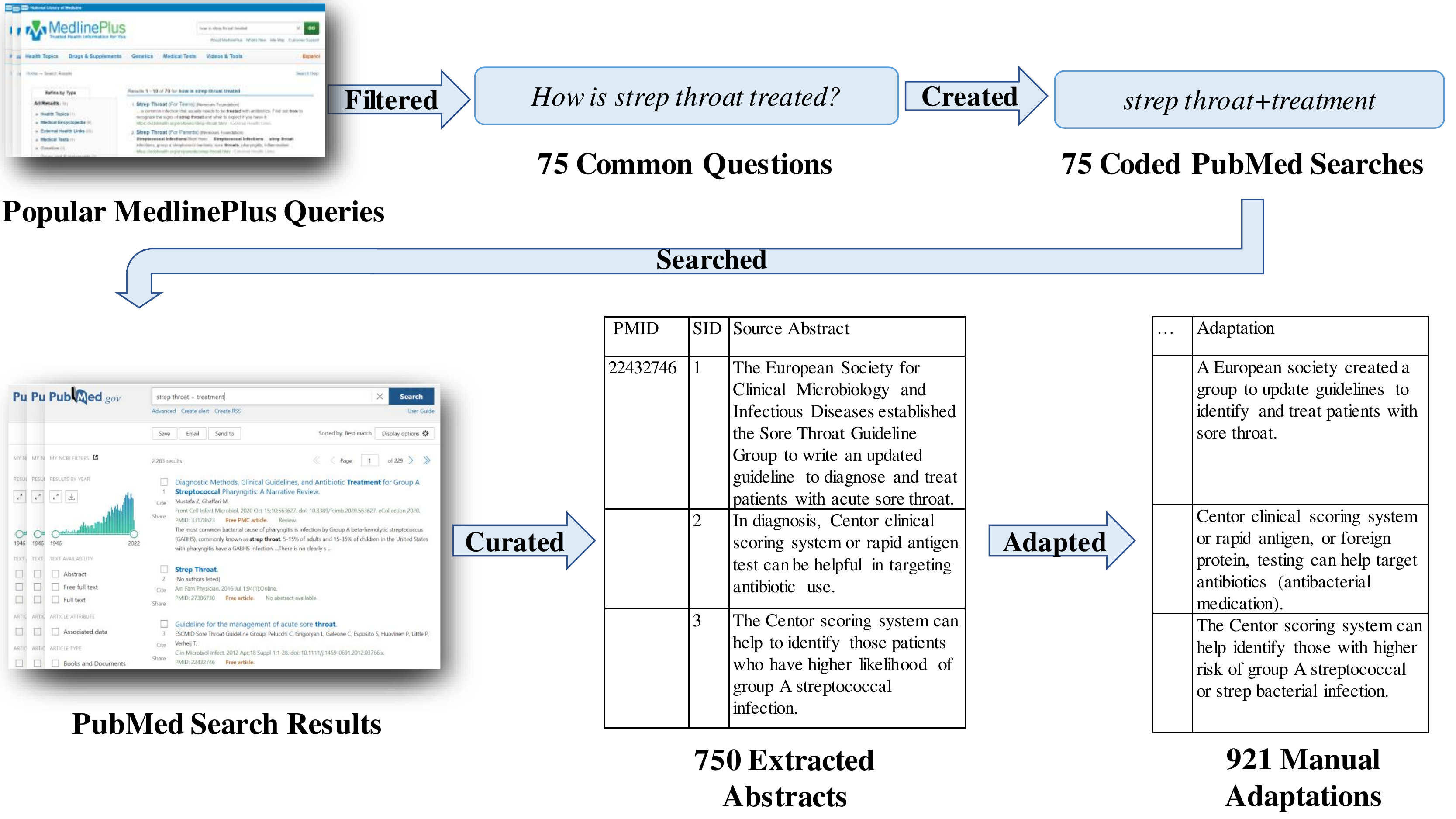}
\caption{Overview representing how questions and PubMed abstracts for the dataset were searched and chosen for annotators to adapt. PMID refers to the PubMed ID from which the example originates from. SID refers to the sentence ID or number of the example sentence from the source abstract.}
\label{fig:1}
\end{figure}

To gather the PubMed abstracts in PLABA, we first filtered questions from MedlinePlus logs based on the frequency of general public queries. Then, a medical informatics expert verified the relevance of and lack of accessible resources to answer each question and chose 75 questions total. For each question, the expert coded its focus (COVID-19, cystic fibrosis, compression devices, etc.) and question type (general information, treatment, prognosis, etc.) to use as keywords in a PubMed search \cite{https://doi.org/10.1002/asi.23806}. Then, the expert selected 10 abstracts from PubMed retrieval results that appropriately addressed the topic of the question, as seen in Figure \ref{fig:1}.

\begin{figure}[ht]
\centering
\includegraphics[width=\linewidth]{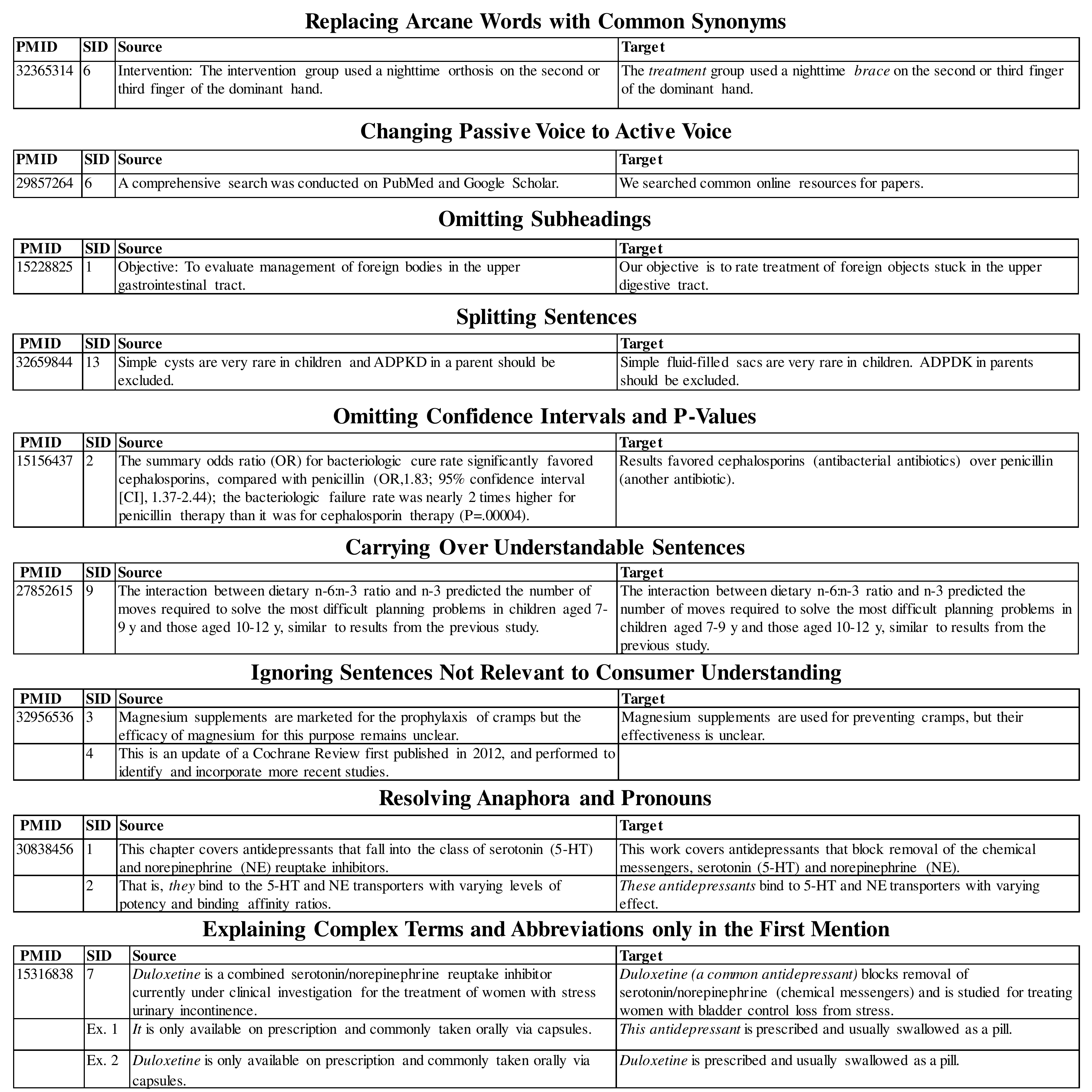}
\caption{Example of the guidelines set for annotators. PMID refers to the PubMed ID from which the example originates from. SID refers to the sentence ID or number of the example sentence from the source abstract. Target refers to the manual adaptation.}
\label{fig:2}
\end{figure}

To create the corresponding adaptations for each abstract in PLABA, medical informatics experts worked with source abstracts separated into individual sentences to create corresponding adaptations across all 75 questions. 
Adaptation guidelines allowed annotators to split long source sentences and ignore source sentences that were not relevant to the general public. Each source sentence corresponds to no, one, or multiple sentences in the adaptation. Creating these adaptations involved syntactic, lexical and semantic simplifications, which were developed in the context of the entire abstract. Examples taken from the dataset can be seen in Table \ref{tab:1}. Specific examples of adaptation guidelines are demonstrated in Figure \ref{fig:2} and included:
\begin{itemize}
    \item Replacing arcane words like "orthosis" with common synonyms like "brace"
    \item Changing sentence structure from passive voice to active voice
    \item Omitting or incorporating subheadings at the beginning of sentences (e.g., "Aim:", "Purpose:")
    \item Splitting long, complex sentences into shorter, simpler sentences
    \item Omitting confidence intervals and other statistical values
    \item Carrying over understandable sentences from the source with no changes into the adaptation
    \item Ignoring sentences that are not relevant to a patient's understanding of the text
    \item Resolving anaphora and pronouns with specific nouns
    \item Explaining complex terms and abbreviations with explanatory clauses when first mentioned
\end{itemize}

\section*{Data Records}

We archived the dataset with Open Science Framework (OSF) at \hyperlink{https://osf.io/rnpmf/}{https://osf.io/rnpmf/}. The dataset is saved in JSON format and organized or "keyed" by question ID. Each key is a question ID that contains a corresponding nested JSON object. This nested object contains the actual question and contains the abstracts and corresponding human adaptations grouped by the PubMed ID of the abstract. Table \ref{tab:2} shows statistics of the abstracts and adaptations. Additional details regarding the data structure can be found in the README file in the OSF archive. 

\section*{Technical Validation}

We measured the level of complexity, the ability to train tools and how well the main points are preserved in the automatic adaptations trained on our data. We first introduce the metrics we used to measure text complexity followed by the metrics to measure text similarity and inter-annotator agreement between manually created adaptations. We use the same text similarity metrics to also compare automatically created adaptations to both the source abstracts and manually created adaptations.

\subsection*{Evaluation metrics}  

To measure text readability and compare the abstracts and manually created adaptations, we use the Flesch-Kincaid Grade Level (FKGL) test\cite{flesch_new_1948}. FKGL uses the average number of syllables per word and the average number of words per sentence to calculate the score. A higher FKGL score for a text indicates a higher reading comprehension level needed to understand the text. 

In addition, we use BLEU\cite{papineni_bleu_2002}, ROUGE\cite{lin_rouge_2004}, and SARI\cite{xu_optimizing_2016,sun_document-level_2021}, commonly used text similarity and simplification metrics, to measure inter-annotator agreement, compare abstracts to manually created adaptations, and evaluate the automatically created adaptations. BLEU and ROUGE look at spans of contiguous words (referred to as n-grams in Natural Language Processing or NLP) to evaluate a candidate adaptation against a reference adaptation. For instance, BLEU-4 measures how many of the contiguous sequences from one to four words in length in the candidate adaptation appear in the reference adaptation. However, BLEU is a measure of precision and penalizes candidates for adding incorrect n-grams. ROUGE is a measure of recall and penalizes candidate adaptations for missing n-grams. Since neither BLEU nor ROUGE is specifically designed for simplification, we also use SARI, which also incorporates the source sentence in order to weight the various operations involved in simplification.  While n-grams are still used, SARI balances (1) addition operations, in which n-grams of the candidate adaptation are shared with the reference adaptation but not the source, (2) deletion operations, in which n-grams appear in the source but neither the reference nor candidate, and (3) keep operations, in which n-grams are shared by all three. We report BLEU-4, ROUGE-1, ROUGE-2, ROUGE-L (which measures the longest shared sub-sequence between a candidate and reference), and SARI. All metrics can account for multiple possible reference adaptations.

\subsection*{Text readability}

To verify that the human generated adaptations simplify the source abstracts, we calculated the FKGL readability scores for both the adaptations and abstracts.  FKGL scores were lower for the adaptations compared to the abstracts (\emph{p} < 0.0001, Kendall’s tau). It is important to note that FKGL does not measure similarity or content preservation, so additional metrics like BLEU, ROUGE, and SARI are needed to address this concern.

\subsection*{Inter-annotator agreement} 

To measure inter-annotator agreement, we used adaptions from the most experienced annotator (who also helped define the guidelines) as reference adaptations. Agreement was measured for all abstracts that were adapted by this annotator and another annotator.
For the inter-annotator agreement metrics of ROUGE-1, ROUGE-2, ROUGE-L, and BLEU-4, the values ranged from 0.4025-0.5801, 0.1267-0.2983, 0.2591-0.4689, and 0.0680-0.2410, respectively, for all adaptations that were done by the reference annotator and another annotator.
As the ROUGE-1 results show, the other annotators included, on average, about half of the words that the reference annotator used. 
As expected, ROUGE-2 values are lower, on average, because as n-grams increase in n, there will be less similarity between adaptations since individuals may use different combinations of words when creating new text.  

We also calculated the similarity between human adaptations and the source abstracts. Using the abstracts as candidates and adaptations as references since BLEU-4 can only match multiple references to a single candidate and not vice versa, the scores in Table \ref{tab:3} show the adaptations contain over half of the same words and a third of the same bi-grams as the source abstracts.  

\begin{figure}[ht]
\centering
\includegraphics[width=\linewidth]{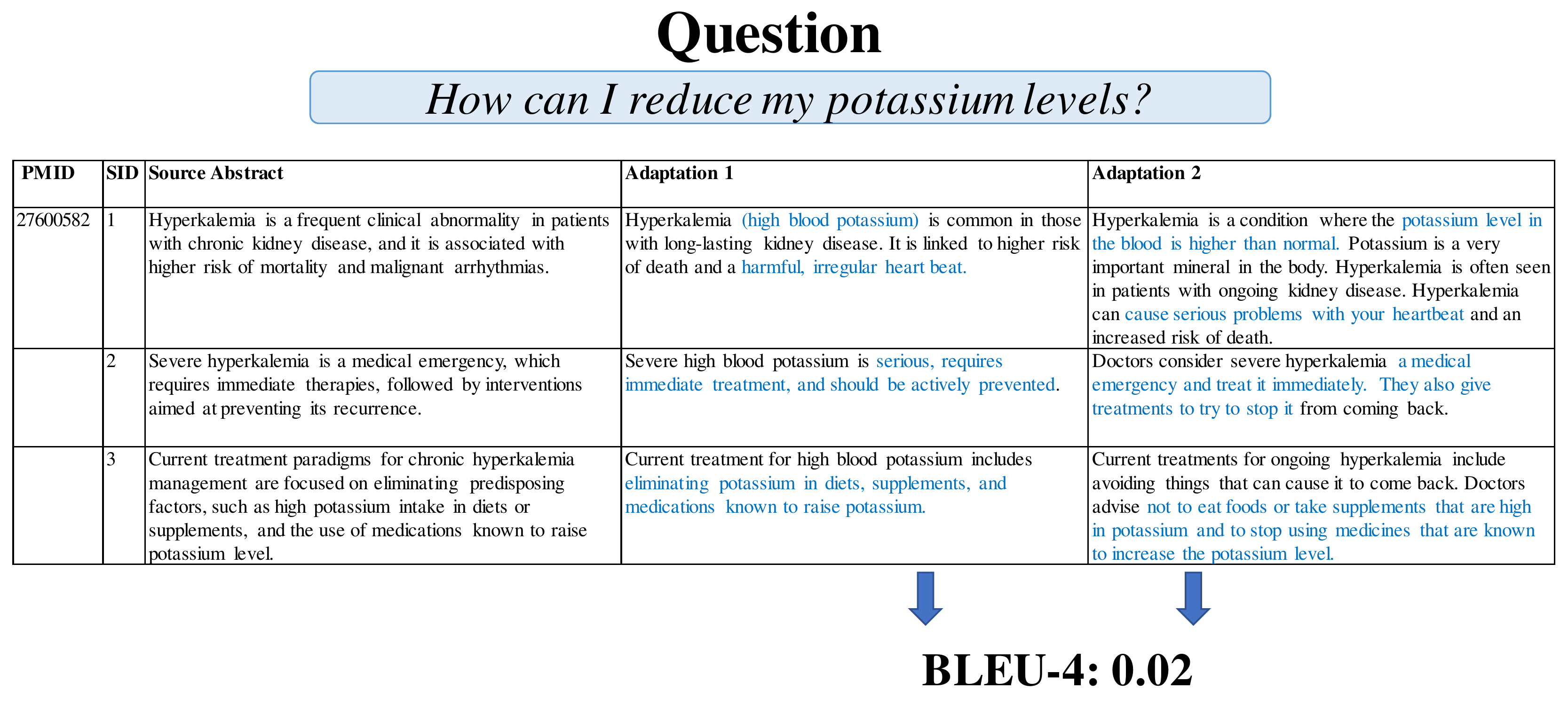}
\caption{Example of the low BLEU-4 score between human adaptations from two different annotators created from the same source abstract and answering the same question. PMID refers to the PubMed ID from which the example originates from. SID refers to the sentence ID or number of the example sentence from the source abstract. Colored text in an adaptation represents parts of the adaptation that strongly differ from the other adaptation.}
\label{fig:3}
\end{figure}

While ROUGE and BLEU are metrics for text similarity, they do not necessarily measure correctness. Even if a pair of adaptations have a low ROUGE or BLEU score, both could be accurate restatements of the source abstract as seen in Figure \ref{fig:3}. While the BLEU-4 score can be low, both adaptations can relevantly describe the topic in response to the example question. The differences between the adaptations can be attributed to synonyms and differences in explanatory content. While BLEU and ROUGE are useful for measuring lexical similarity, calculating differences between adaptations like these is more nuanced. To address this issue, researchers are actively developing new metrics\cite{kryscinski_evaluating_2020}. 

\subsection*{Experimental benchmarking} 

To benchmark the PLABA dataset and show its use in evaluating automatically generated adaptations, we used a variety of state-of-the-art deep learning algorithms listed below: 

\subsubsection*{Text-to-Text Transfer Transformer (T5)} 

T5\cite{raffel_exploring_2020} is a transformer-based\cite{vaswani2017attention} encoder-decoder model with a bidirectional encoder setup similar to BERT\cite{devlin_bert_2019} and an autoregressive decoder that is similar to the encoder except with a standard attention mechanism. Instead of training the model on a single task, T5 is pre-trained on a vast amount of data and on many unsupervised and supervised objectives, including token and span masking, classification, reading comprehension, translation, and summarization. The common feature of every objective is that the task can be treated as a language-generation task, in which the model learns to generate the proper textual output in response to the textual prompt included in the input sequence. As with other models, pre-training has been shown to achieve state-of-the-art results on many NLP tasks\cite{kryscinski_evaluating_2020}\cite{raffel_exploring_2020}\cite{radford2019language}. In our experiments, we use the T5-Base model. 

\subsubsection*{Pre-training with Extracted Gap-sentences for Abstractive SUmmarization Sequence-to-sequence (PEGASUS)}

PEGASUS\cite{zhang_pegasus_2020} is another transformer-based encoder-decoder model; however, unlike T5, PEGASUS is pre-trained on a unique self-supervised objective. With this objective, entire sentences are masked from a document and collected as the output sequence for the remaining sentences of the document. In other words, PEGASUS is designed for abstractive summarization and similar tasks, achieving human performance on multiple datasets. In our experiments, we use the PEGASUS-Large model. 

\subsubsection*{Bidirectional autoregressive transformer (BART)}

BART\cite{lewis_bart_2020} is another transformer-based encoder-decoder that is pre-trained with a different objective. Instead of training the model directly on data with a text-to-text objective or summarization-specific objective, BART was pre-trained on tasks such as token deletion and masking, text-infilling, and sentence permutation. These tasks were developed to improve the model’s ability to understand the content of text before summarizing or translating it. After this pre-training, BART can be fine-tuned for downstream tasks of summarization or translation with a more specific dataset to output higher quality text. These datasets include the CNN Daily Mail\cite{nallapati_abstractive_2016} dataset, a large news article dataset designed for summarization tasks. In our experiments, we use the BART-Base model and BART-Large model fine-tuned on the CNN Daily Mail dataset (BART-Large-CNN).

\subsubsection*{Experimental Setup} 
\begin{figure}[ht]
\centering
\includegraphics[width=.7\textwidth]{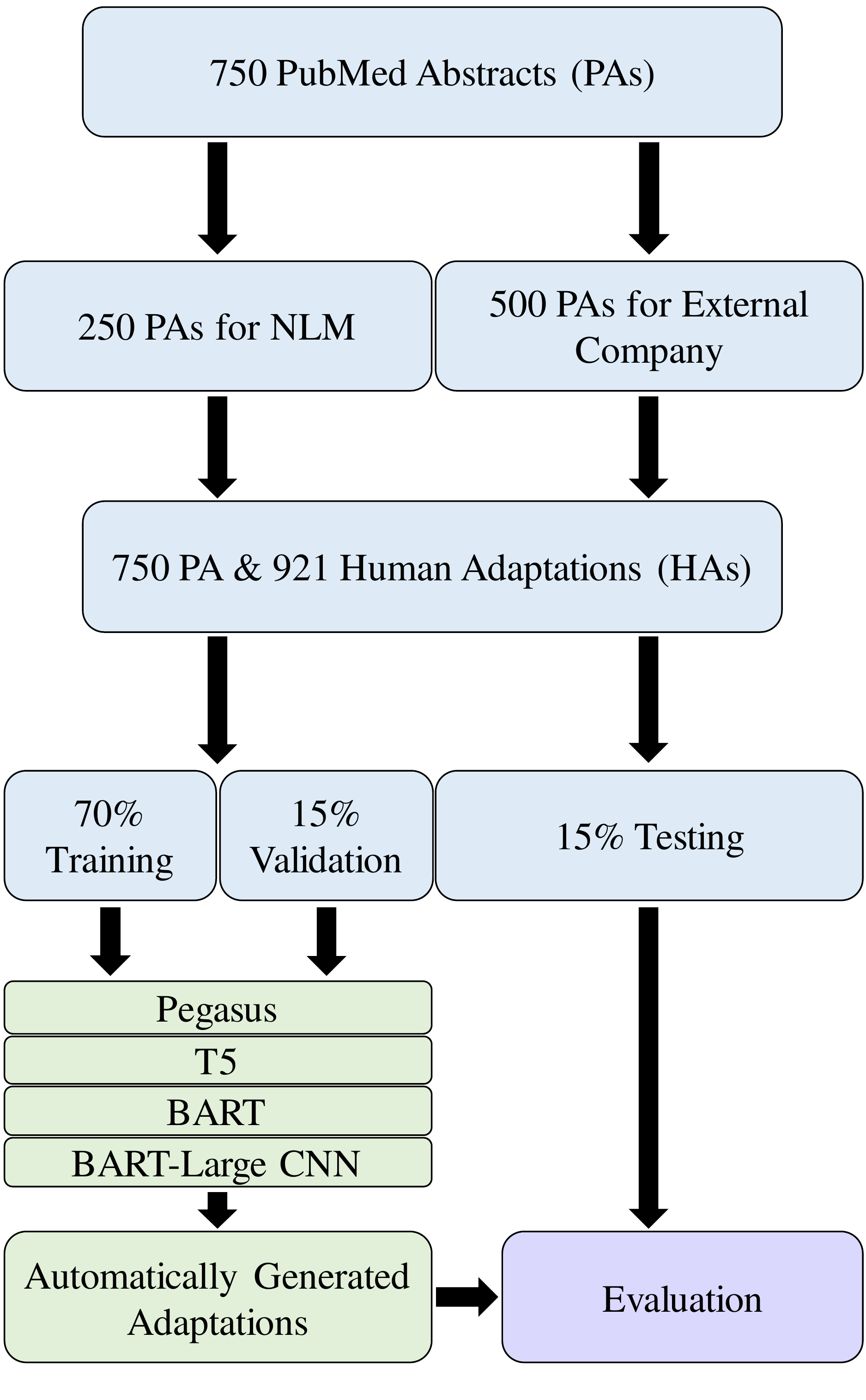}
\caption{Overview representing how PubMed abstracts and human adaptations are split for training and testing models.}
\label{fig:4}
\end{figure}

For our experiments, all deep learning models were trained using the abstracts and adaptations in the PLABA dataset. Each PubMed abstract is used as the source document, and the human generated adaptations are used as the references. The dataset was divided such that 70\% was used for training, 15\% for validation, and 15\% for testing. We utilized the pre-trained models from Hugging Face\cite{wolf_transformers_2020}, and each model was trained with the AdamW optimizer and the default learning rate of 5e-5 for 20 epochs using V100X GPUs (32 GB VRAM) on a shared cluster. Maximum input sequence length was set to 512 tokens except for the BART models, in which the maximum was set to 1024. Validation loss was measured every epoch, and the checkpoint model with the lowest validation loss was used for test set evaluation. An overview of the experiments can be seen in Figure \ref{fig:4}

\subsubsection*{Results} 
Table \ref{tab:4} shows the FKGL scores between the automatically generated adaptations, all of which were significantly lower than the abstracts and significantly higher than the manually crafted adaptations (\emph{p} < 0.0001, Kendall’s tau). Table \ref{tab:5} shows the comparison between the automatically generated adaptations and the human generated adaptations with ROUGE and BLEU and the comparison between the automatically generated adaptations, human generated adaptations, and source abstracts with SARI. Table \ref{tab:6} shows the comparison between the automatically generated adaptations and the source abstracts with ROUGE and BLEU. 
It is interesting to note that the automatically generated adaptations are more readable than the abstracts but less readable than the human generated adaptations. Thus, the dataset gives the models sufficient training data to develop outputs that outperform the source abstracts in terms of readability. Regarding SARI, the models tend to perform comparably in terms of simplification. In terms of ROUGE and BLEU, the automatically generated adaptations tend to share more n-grams with the source abstracts rather than the human generated adaptations. This relationship is potentially because the abstracts tend to be shorter than the adaptations, as seen in Table \ref{tab:2}. This may make it easier for the automatically generated adaptations to share more contiguous word sequences with the abstracts relative to the human generated adaptations. In addition, the choice of metrics used for evaluation will influence the reported performance of a model. 

An example of the automatically generated adaptations from each model in response to the same abstract is shown in Table \ref{tab:7}. These demonstrate that the PLABA dataset, in addition to being a high-quality test set, is useful for training generative deep learning models with the objective of text adaptation of scientific articles. Since there are no existing manually crafted datasets for this objective, PLABA can be a valuable dataset for benchmarking future research in this domain.

\section*{Usage Notes}

We have added instructions in the README file of our OSF repository that show how to use the PLABA dataset. Pre-processing the dataset and evaluating adaptation algorithms on it can be located in the code scripts at our GitHub repository given below. 

\section*{Code availability}

Code scripts to pre-process PLABA and reproduce the benchmark results of the experiments can be found at \hyperlink{https://github.com/attal-kush/PLABA}{https://github.com/attal-kush/PLABA}.

\bibliography{main}

\section*{Acknowledgements}

This work was supported by the Intramural Research Program of the National Library of Medicine (NLM), National Institutes of Health, and utilized the computational resources of the NIH HPC Biowulf cluster (http://hpc.nih.gov).

\section*{Author contributions statement}

K.A. created the code scripts for data pre-processing and deep learning experiments, contributed to adaptation guidelines, contributed to creating the manual adaptations, and wrote and edited the manuscript.  
B.O. contributed to adaptation guidelines and edited the manuscript. 
D.D.-F conceived the project, edited the manuscript, and provided feedback at all stages of the study.  

\section*{Competing interests}

The authors declare no competing interests. 

\section*{Figures \& Tables}

\begin{table}[ht]
\centering
\begin{tabular}{|l|p{0.8\linewidth}|}
\hline
\textbf{Data Type} & \textbf{Text} \\
\hline
Question & \emph{How is strep throat treated?} \\
\hline
Abstract & Most patients who seek medical attention for sore throat are concerned about streptococcal tonsillopharyngitis, but fewer than 10\% of adults and 30\% of children actually have a streptococcal infection. Group A beta-hemolytic streptococci (GAS) are most often responsible for bacterial tonsillopharyngitis, although Neisseria gonorrhea, Arcanobacterium haemolyticum (formerly Corynebacterium haemolyticum), Chlamydia pneumoniae (TWAR agent), and Mycoplasma pneumoniae have also been suggested as possible, infrequent, sporadic [...]\\
\hline
Adaptation & Most people who go to a doctor for sore throat are worried they have a strep throat and tonsil infection, but fewer than 10\% of adults and 30\% of children actually have a strep infection. Group A strep bacteria are the most common cause of bacterial strep throat and tonsil infection, but other bacteria known to cause sexually-transmitted gonorrhea or chlamydia, or head, neck, and lung infections occasionally might [...]\\
\hline
Question & \emph{How do you assist an unconscious victim who is already vomiting?} \\
\hline
Abstract & The tongue is the most common cause of upper airway obstruction, a situation seen most often in patients who are comatose or who have suffered cardiopulmonary arrest. Other common causes of upper airway obstruction include edema of the oropharynx and larynx, trauma, foreign body, and [...] \\
\hline
Adaptation & The tongue is the most common cause of blocked upper airways, seen most often in people in comas or cardiac arrest (abrupt heart stop). Other common causes of blocked upper airways include swelling of the middle part of the throat and voice box, injury, objects that shouldn’t be swallowed, and [...] \\
\hline
\end{tabular}
\caption{\label{tab:1}Examples of questions, abstracts, and adaptations in PLABA.}
\end{table}

\begin{table}[ht]
\centering
\begin{tabular}{|l|l|l|l|l|l|}
\hline
\textbf{Data Type} & \textbf{Count} & \multicolumn{2}{l|}{\textbf{Words}} & \multicolumn{2}{l|}{\textbf{Sentences}} \\
\cline{3-6}
& & \textbf{Average} & \textbf{S.d.} & \textbf{Average} & \textbf{S.d.} \\
\hline
Questions & 75 & 10 & 6 & 1 & 0\\
\hline
Abstracts & 750 & 240 & 95 & 10 & 4\\
\hline
Adaptations & 921 & 244 & 95 & 12 & 5\\
\hline
\end{tabular}
\caption{\label{tab:2}Average number of words and sentences per data type.}
\end{table}


\begin{table}[ht]
\centering
\begin{tabular}{|l|l|l|l|l|}
\hline
\textbf{Comparison Type} & \textbf{ROUGE-1} & \textbf{ROUGE-2} & \textbf{ROUGE-L} & \textbf{BLEU-4}\\
\hline
Adaptations vs Abstracts & 0.58 & 0.36 & 0.50 & 0.39\\
\hline
\end{tabular}
\caption{\label{tab:3}ROUGE-1, ROUGE-2, ROUGE-L, and BLEU-4 using human adaptations as references and abstracts as candidates.}
\end{table}

\begin{table}[ht]
\centering
\begin{tabular}{|l|l|l|}
\hline
\textbf{Data Type} & \multicolumn{2}{l|}{\textbf{FKGL}}\\
\cline{2-3}
& \textbf{Average} & \textbf{S.d.} \\
\hline
Abstracts & 15.78 & 8.06\\
\hline
Adaptations & 12.04 & 2.39\\
\hline
T5 & 13.83 & 2.93\\
\hline
PEGASUS & 13.84 & 2.80\\
\hline
BART-Base & 13.37 & 2.59\\
\hline
BART-Large-CNN & 13.19 & 2.66\\
\hline
\end{tabular}
\caption{\label{tab:4}FKGL scores for automatically generated adaptations.}
\end{table}

\begin{table}[ht]
\centering
\begin{tabular}{|l|l|l|l|l|l|}
\hline
\textbf{Algorithm Type} & \textbf{ROUGE-1} & \textbf{ROUGE-2} & \textbf{ROUGE-L} & \textbf{BLEU-4} & \textbf{SARI}\\
\hline
T5 & 0.56 & 0.30 & 0.42 & 0.28 & 0.34\\
\hline
PEGASUS & 0.57 & 0.31 & 0.43 & 0.30 & 0.32\\
\hline
BART-Base & 0.56 & 0.31 & 0.43 & 0.29 & 0.33\\
\hline
BART-Large-CNN & 0.58 & 0.31 & 0.42 & 0.29 & 0.26\\
\hline
\end{tabular}
\caption{\label{tab:5}Automatically generated adaptations compared to human adaptations and (only for SARI) source abstracts.}
\end{table}

\begin{table}[ht]
\centering
\begin{tabular}{|l|l|l|l|l|}
\hline
\textbf{Algorithm Type} & \textbf{ROUGE-1} & \textbf{ROUGE-2} & \textbf{ROUGE-L} & \textbf{BLEU-4}\\
\hline
T5 & 0.77 & 0.62 & 0.74 & 0.55 \\
\hline
PEGASUS & 0.79 & 0.64 & 0.74 & 0.60 \\
\hline
BART-Base & 0.76 & 0.62 & 0.73 & 0.57 \\
\hline
BART-Large-CNN & 0.73 & 0.56 & 0.69 & 0.50 \\
\hline
\end{tabular}
\caption{\label{tab:6}Automatically generated adaptations compared to source abstracts.}
\end{table}

\begin{table}[ht]
\centering
\begin{tabular}{|l|p{0.8\linewidth}|}
\hline
\textbf{Algorithm Type} & \textbf{Generated Text} \\
\hline
Question & \emph{How to treat a bakers cyst?} \\
\hline
PEGASUS & Osteoarthritis of the knee is a common disease, often associated with a Baker's cyst. Besides osteoarthritis, also other joint diseases (diseases that cause pain and swelling in the joints) can be [...] A decrease in volume more than 25\% compared to the baseline could be achieved for 75\% of patients. Most patients respond to treatment. Radiation therapy of knee osteoarthritis (damage to cartilage and bone) is effective that decreases the number of affected knees. However, whether radiotherapy is the best treatment is still being studied.\\
\hline
T5 & Osteoarthritis of the knee (knee arthritis) is a common disease. It is often associated with the Baker's cyst (a cyst in the back of your knee). Besides arthritis, also other joint diseases of knees [...] A decrease of more than 25\% compared to the baseline could be achieved for 75\% of patients in short term and 79\% of people in long term follow up. Conclusion: Radiation therapy is an efficient treatment that decreases the amount of swollen area. Most patients respond to this treatment. Whether radiation is effective for Baker'cyst without associated arthritis must be further examined.\\
\hline
BART-Base & Osteoarthritis of the knee is a common disease, often associated with a Baker's cyst (a small, bone-related lump). Also, other joints in the leg can be causative [...] A decrease of more than 25\% compared to the baseline could be achieved for 75\% of patients at the beginning and 79\% (long-lasting) after the treatment. This conclusion is that radiation therapy of knee osteosarcoma is an important treatment that decreases the amount and severity of an osteoporosis cysteostasis. Most patients respond to it. However, it is still unclear how effective this treatment is.\\
\hline
BART-Large-CNN & Osteoarthritis of the knee is a common disease, often associated with a Baker's cyst, a growth of swollen tissues in the joints. Besides bone pain, other problems with the ligament that runs between the thigh bone and knee joint can cause [...] A decrease in volume of more than 25\% compared to the baseline could be achieved for 75\% of patients in both the short term and the long term. In conclusion, radiation therapy of knee bone arthritis is an helpful treatment that decreases the size of an acchly Cyst. Most patients respond to treatment. Whether radiation is the best treatment to treat a Bakty for a kyst without the associated pain caused by bone damage is further examined.\\
\hline
Question & \emph{Why is transferrin and iron low in covid patients but ferritin high?} \\
\hline
PEGASUS & Ferritin, which includes twenty-four light and heavy chains in varying proportions in different tissues, is primarily responsible for maintaining the body's iron metabolism. [...] It can be combined with monoclonal antibodies, antioxidants, corticosteroids (antibodies that fight inflammation) and lactoferrin (a type of protein that protects cells from infection). In this article, we analyze the antiviral and antifibrotic activity of iron-chelator drugs, thereby promoting iron depletion therapy as a potentially innovative treatment strategy for Covid-19 (chronic kidney disease).\\
\hline
T5 & Ferritin, which includes twenty-four light and heavy chains in varying proportions in different tissues, is primarily responsible for maintaining the body's iron metabolism. [...] This can be combined with monoclonal antibodies, antioxidants, corticosteroids (glucosides) and lactoferrin to make iron therapy effective for covid-19 victims. In this article, we analyze the antiviral and antifibrotic activity of iron agents, thereby promoting iron depletion therapy as potentially innovative treatment strategy for the covd-19.\\
\hline
BART-Base & Ferritin, which includes twenty-four light and heavy chains in varying proportions in different tissues, is primarily responsible for maintaining the body's iron metabolism. [...] These drugs can be combined with monoclonal antibodies, antioxidants, corticosteroids (antibody drugs that help fight diseases that attack the brain and spinal cord) to make iron-chelation therapy effective for Covid-18 victims. In this article, we analyze the antiviral and antifibrotic (anti-seizure) activity of these drugs, thereby promoting iron depletion therapy as a potentially innovative treatment strategy for COV-17.\\
\hline
BART-Large-CNN & Ferritin, which includes twenty-four light and heavy chains in varying proportions in different tissues, is mainly responsible for maintaining the body's iron metabolism. [...] It can be combined with antibodies, antioxidants, corticosteroids (drugs that fight inflammation), and lactoferrin to make iron removal therapy effective for patients. In this article, we analyze the antiviral and antifibrotic (redness and swelling from infection-fighting) activity of iron chylides, thereby promoting iron reduction therapy as a potentially new treatment strategy for Covid-9.\\
\hline
\end{tabular}
\caption{\label{tab:7}Examples of adaptations created by PEGASUS, T5, BART-Base, BART-Large-CNN.}
\end{table}

\end{document}